\documentclass[sigconf]{acmart}
\usepackage{mathtools}
\AtBeginDocument{%
  }

\copyrightyear{2026}
\acmYear{2026}
\setcopyright{cc}
\setcctype{by}
\acmConference[CAIN '26]{2026 IEEE/ACM 5th International Conference on AI Engineering - Software Engineering for AI}{April 12--13, 2026}{Rio de Janeiro, Brazil}
\acmBooktitle{2026 IEEE/ACM 5th International Conference on AI Engineering - Software Engineering for AI (CAIN '26), April 12--13, 2026, Rio de Janeiro, Brazil}
\acmPrice{}
\acmDOI{10.1145/3793653.3793764}
\acmISBN{979-8-4007-2475-6/2026/04}

\begin{document}

\title{SETA: Statistical Fault Attribution for Compound AI Systems}

\author{Sayak Chowdhury}
\email{sayak.chowdhury@iiitb.ac.in}
\orcid{0009-0006-0191-7410}
\affiliation{
  \institution{International Institute of Information Technology Bangalore}
  \city{Bengaluru}
  \state{Karnataka}
  \country{India}
}

\author{Meenakshi D'Souza}
\orcid{0000-0003-3640-6240}
\affiliation{
  \institution{International Institute of Information Technology Bangalore}
  \city{Bengaluru}
  \state{Karnataka}
  \country{India}
}

\renewcommand{\shortauthors}{Chowdhury and D'Souza}

\begin{abstract}
  Modern AI systems increasingly comprise multiple interconnected neural networks to tackle complex inference tasks.
  Testing such systems for robustness and safety entails significant challenges.
  Current state-of-the-art robustness testing techniques, whether black-box or white-box, have been proposed and implemented for single-network models and do not scale well to multi-network pipelines.
  We propose a modular robustness testing framework that applies a given set of perturbations to test data.
  Our testing framework supports (1) a component-wise system analysis to isolate errors and (2) reasoning about error propagation across the neural network modules.
  The testing framework is architecture and modality agnostic and can be applied across domains.
  We apply the framework to a real-world autonomous rail inspection system composed of multiple deep networks and successfully demonstrate how our approach enables fine-grained robustness analysis beyond conventional end-to-end metrics.
\end{abstract}



\received{16 October 2025}
\received[revised]{23 October 2025}
\received[accepted]{5 January 2026}

\maketitle

\section{Introduction}
\label{sec:introduction}

Modern AI systems are increasingly architected as compound AI systems~\cite{compound-ai-blog} comprising of specialized models, retrievers, and software components composed into complex pipelines to achieve state-of-the-art results, instead of large, monolithic models.
This compositional approach is now standard even in safety-critical domains, such as perception and planning stacks in autonomous vehicles and multi-modal diagnostic tools in healthcare.
While modularity aids development, it amplifies challenges associated with debugging and ensuring system safety against the problem of cascading failures thereby introducing the risk of emergent misalignment.
Small, localized faults in any upstream component such as minor data drifts, model miscalibrations, or unexpected edge cases, can propagate and proliferate through the pipeline resulting in catastrophic system-level safety violations.

Debugging deep learning pipelines remains challenging~\cite{MLtestingSurvey}, as conventional robustness techniques including adversarial attacks like Fast Gradient Signed Method~\cite{szegedy_goodfellow_fgsm} and Projected Gradient Descent~\cite{PGDmadry} primarily target end-to-end behaviour or operate at the input/output level of individual networks.

Formal verification frameworks (e.g., Reluplex~\cite{reluplex}, ERAN~\cite{singh_abstract_2019}) typically analyse isolated DNNs under bounded-input assumptions, offering no support for fault localization within a composed pipeline. 
Consequently, current methods cannot answer questions such as: Which subnetwork in the pipeline is most susceptible to perturbations? or How do small internal errors propagate through a sequence of models?
Coverage-based testing tools like DeepXplore~\cite{deepxplore} introduce neuron coverage metrics to guide test input generation, while DeepTest~\cite{deeptest} employs image-level transformations (e.g., rain, blur) to expose failures in self-driving systems.
CLEVER~\cite{clever_2018} and related metrics quantify adversarial robustness via Lipschitz-based scores, but focus on single-network behaviour.
Formal verification tools such as Planet~\cite{ehlers_planet}, Marabou~\cite{marabou,marabou2}, and abstract interpretation-based approaches like Neurify~\cite{neurify} or ERAN provide provable guarantees, but remain restricted to isolated models and lack compositional analysis. Bunel et al.~\cite{bunel_2018} survey these techniques and underscore their scalability limitations for real-world systems.
While traditional verification techniques focus on analysing individual models in isolation, our work takes a complementary approach: we inject untargeted adversarial noise and empirically trace how failures propagate through multi-component pipelines to evaluate the robustness of each module.
The MLOps ecosystem already offers mature tools for managing the AI lifecycle—frameworks like Kedro support modular, maintainable data pipelines; MLflow enables experiment tracking, model versioning, and deployment; and tools like Evidently AI monitor data drift and performance regressions in production.
However, these platforms stop short of fine-grained, \textit{causal fault attribution}.
When a complex AI pipeline fails, existing tools can detect that a failure occurred but cannot explain why or why it happened within the pipeline's execution or perform fine-grained analysis based on formal specifications.

To bridge this critical gap, we introduce \textbf{SETA}, a novel framework that combines \textit{per-component Metamorphic Testing (MT)} with \textit{Execution Trace Analysis}.
SETA defines oracle-free behavioural specifications via metamorphic relations and attributes violations to specific components by analysing the dynamic execution graph generated during inference.
In doing so, it pinpoints the earliest point of failure in the causal chain.
Rather than replacing tools like Kedro or MLflow, SETA complements them by adding a diagnostic layer that enables precise, empirical fault localization in multi-model AI systems.

This work makes the following key contributions:
(1) We propose \textit{SETA}, a modular framework that integrates \textit{Metamorphic Testing} with \textit{Execution Trace Analysis} to localize faults in complex AI pipelines enabling component-wise robustness analysis by empirically tracing how perturbations propagate through dynamically constructed execution graphs.
(2) By defining oracle-free behavioral specifications through metamorphic relations, SETA supports scalable, testable abstractions for black-box models.
(3) Due to its modular design, SETA allows users to define and plug in different classes of metamorphic relations, supporting the exploration of interpretability in otherwise opaque systems.
(4) We demonstrate SETA’s effectiveness in pinpointing failure origins and surfacing hidden vulnerabilities in multi-stage AI pipelines, where traditional verification or monitoring tools fall short.

\section{Background}
\label{sec:background}
The critical diagnostic challenge in testing compound AI systems falls squarely between two testing paradigms: End-to-End (E2E) and Isolated Component Testing (ICT).
E2E testing treats systems as black-boxes and can identify failure occurrences but is unable to perform credit attribution to internal components, while on the other hand, ICT is insufficient, being unable to detect emergent faults arising from dynamic interaction between modules or analyze how errors propagate through the pipeline.
This necessitates the requirement for a testing methodology that can operate \textit{in situ} and is capable of evaluating components within the context of the full system to perform fine-grained fault localization.

\subsection{Metamorphic Testing for pseudo-Oracles}
\label{sec:metamorphic-testing}
A primary obstacle in testing AI components is the difficulty of defining the exact, correct output for any given input in the absence of any annotated ground truth, also known as the \textit{test-oracle problem}~\cite{oracle-test}.
The absence of such a well-specified input-output mapping, particularly for intermediate components of a dynamic pipeline, can lead to the emergence of erroneous deviations that accumulate and propagate unnoticed throughout the system.
{Metamorphic Testing (MT)} is a powerful technique that can be utilized to alleviate this problem~\cite{MR} by verifying preservation of necessary, user-specified properties, known as \textit{Metamorphic Relations (MRs)} across transformations of input data, rather than verifying a single input-output pair for each constituent components.
An MR specifies an expected relationship between a system's outputs for a source input $x$ and a perturbed input $\tilde{x}$ which has undergone a transformation $g(x)$.
A single MR is a predicate over tuples of inputs and outputs that must hold for a correct system such that
$$M(x, \tilde{x}, f(x), f(\tilde{x})) = \texttt{True}$$

For instance, metamorphic relations designed for object detectors may require applying fog synthetically to an image while preserving a subset of the original detections without introducing spurious objects in the output predictions.
Violations of such metamorphic relations signal potential faults.
A library of metamorphic relations can thus be defined (e.g., invariance, subset, equivalence, etc.) based on the properties empirically characterizing the tasks performed by each component and minimum safety specifications associated, to enable robust, pseudo-oracle specifications for the entire system.

\subsubsection{Library of Metamorphic Relations}
\label{sec:library-of-MRs}
The framework utilizes multiple metamorphic relations that have been designed in the context of compound AI systems that address vision-related inference tasks.
The four main vision tasks that are performed are (i) Image Classification, (ii) Object Localization, (iii) Object Detection and (iv) Segmentation.
The following details the metamorphic relations established for each task under consideration and introduces a generalized MR formulation that facilitates the extension of the MR set to incorporate user-defined verification properties.

\paragraph{Generic Metamorphic Relations.}
Each metamorphic relation is defined by wrapping relevant parameters within two well-known functions that check either for inequality or equality constraints.
\textbf{Kronecker delta} $\left(\delta(\theta,\tau) \coloneqq [\theta = \tau]\right)$ performs an equality check for a parameter $\theta$ against the threshold $\tau$ and returns either 1 or 0 based on the whether the condition is met.
\textbf{Heaviside step function } $\left(H(\theta, \tau) \coloneqq [\theta \geq \tau]\right)$, similarly, performs an inequality check for some parameter $\theta$ against corresponding threshold $\tau$.
The domain of both functions spans the entire real line $\mathbb{R}$, thereby enabling a generalized treatment of relational constraints across diverse quantitative measures.
Each metamorphic relation $M$ under this definition will be of the form:
$$
M(x, \tilde{x}; f) \coloneqq
\begin{cases}
\delta(\theta, \tau) \quad \text{equality constraint}\\
H(\theta, \tau) \quad \text{inequality constraint}
\end{cases}
$$
where $M(x, \tilde{x}; f)$ represents the relation over a component $f$.
This representation naturally extends to abstract intervals, i.e., regions of permissible variability around a nominal value.
By replacing the scalar parameter–threshold pair $(\theta, \tau)$ with higher-order subroutines that estimate admissible value ranges, the formulation can encode complex relational properties such as bounded perturbation behaviour or approximate equivalence under transformation.
Consequently, each MR can be composed together, as shown in section~\ref{sec:composite-metamorphic-relations}, to define constraints over an abstract interval in $\mathbb{R}^n$, where the satisfaction of the relation corresponds to the inclusion of the observed metric within that interval.

\textbf{Image Classification.}
Components that perform classification tasks take inputs and return labels or a distribution of values across labels.
The component outputs should remain invariant under spatial transformations like translations, rotations and should not deviate significantly under the presence of additive noise.
For cases where only the output \textit{label} is accessible, the metamorphic relation can be defined as $M_\textit{label}(x, \tilde{x}; f) \coloneqq \delta(\theta, \tau)$, where $\theta$, $\tau$ are the predicted and actual labels respectively.
Similarly when a vector of distributions is provided by the component, the metamorphic relation would be $M_\textit{distb}(x, \tilde{x}; f) \coloneqq H(\theta, \tau)$, where $\theta = \left\lVert f(x) - f(\tilde{x}) \right\rVert_\infty$ is the $L_\infty$-norm difference between the actual and predicted labels and $\tau$ is the maximum $L_\infty$ threshold that can be tolerated.

\textbf{Object Localization.}
In addition to object labels coordinate values are provided via regression in the form of $(c_x, c_y, c_w, c_h)$ tuples where $c_x,c_y$ are the lower-left $c_x$ and $c_y$ coordinates of a bounding box and $c_w, c_h$ are the width, height respectively.
For class labels, relations identical to those defined for image classification task are used.
For localization, specific constraints are defined to verify the degree of spatial overlap between the predicted and ground-truth bounding boxes.
The overlap ratio must exceed a predefined threshold $\tau$, and the resulting relation is $M_\textit{IoU}(x, \tilde{x}; f) \coloneqq H(\theta, \tau)$, where the \textit{Intersection over Union} between the two bounding box coordinates is $\theta = \frac{A_I}{A_U}$.
Here, $A_I = \max \left(0, \min \left( c_x + c_w, c_{\tilde{x}} + c_{\tilde{w}} \right) - \max \left( c_x, c_{\tilde{x}} \right) \right) \cdot \max \left( 0, \min\left(c_y + c_h, c_{\tilde{y}} + c_{\tilde{h}}\right) - \max\left(c_y, c_{\tilde{h}} \right) \right)$
and $A_U = \left(c_w \cdot c_h\right) + \left(c_{\tilde{w}}\cdot c_{\tilde{h}}\right) - A_I$ represent the areas of Intersection $A_I$ and Union $A_U$.

\textbf{Object Detection.}
It extends classification and localization of objects to any $k$ number of objects within the same input image.
The constraints for this task are a combination of those defined for Image Classification and Object Localization only extended to a set of $k$ objects.
The output returned by an object detection component is defined as a dynamic set of $k$ tuples $f(x) = \left\{D^{(i)}\mid i = 1, \dots, k\right\}$ with each tuple of the form $\left(c_x^{(i)}, c_y^{(i)}, c_w^{(i)}, c_h^{(i)}, c^{(i)}\right)$, such that, $c_x^{(i)}, c_y^{(i)}, c_w^{(i)}$ and $c_h^{(i)}$ are the values corresponding to the $i$-th object and $c^{(i)}$ is the predicted class or label distribution~\cite{NNVobj,elboher_formal_2024}.
Thus, for every object detected by the model there are two constraints $M_\textit{IoU}(x, \tilde{x}; f)$ and, $M_\textit{label}(x, \tilde{x}; f)$ or $M_\textit{distb}(x, \tilde{x}; f)$.

\textbf{Image Segmentation.}
Similar to Object Detection an input image may contain up to $k$ different semantic objects. However, instead of determining localization and class labels for each object, every pixel in the input image is assigned a semantic label.
Therefore, in any image of dimensions $h \times w$, there will be $h \times w$ pixel-level predictions made, each prediction either a single categorical label or probability distribution over $k$ semantic classes.
Formally, for an input image $x \in \mathbb{R}^{h\times w\times nc}$ (with $nc$ denoting the number of input channels), the segmentation model outputs $f(x) \in \mathbb{R}^{h\times w\times k}$ which can be decomposed into $k$ binary masks $\{B_1, B_2, \dots, B_k\}$, where each mask $B_i \in \{0,1\}^{h\times w}$ encodes the pixel-wise prediction for the $i$-th semantic class.
\begin{figure}[htbp]
    \centering
    \begin{subfigure}{0.15\textwidth}
        \includegraphics[width=\linewidth]{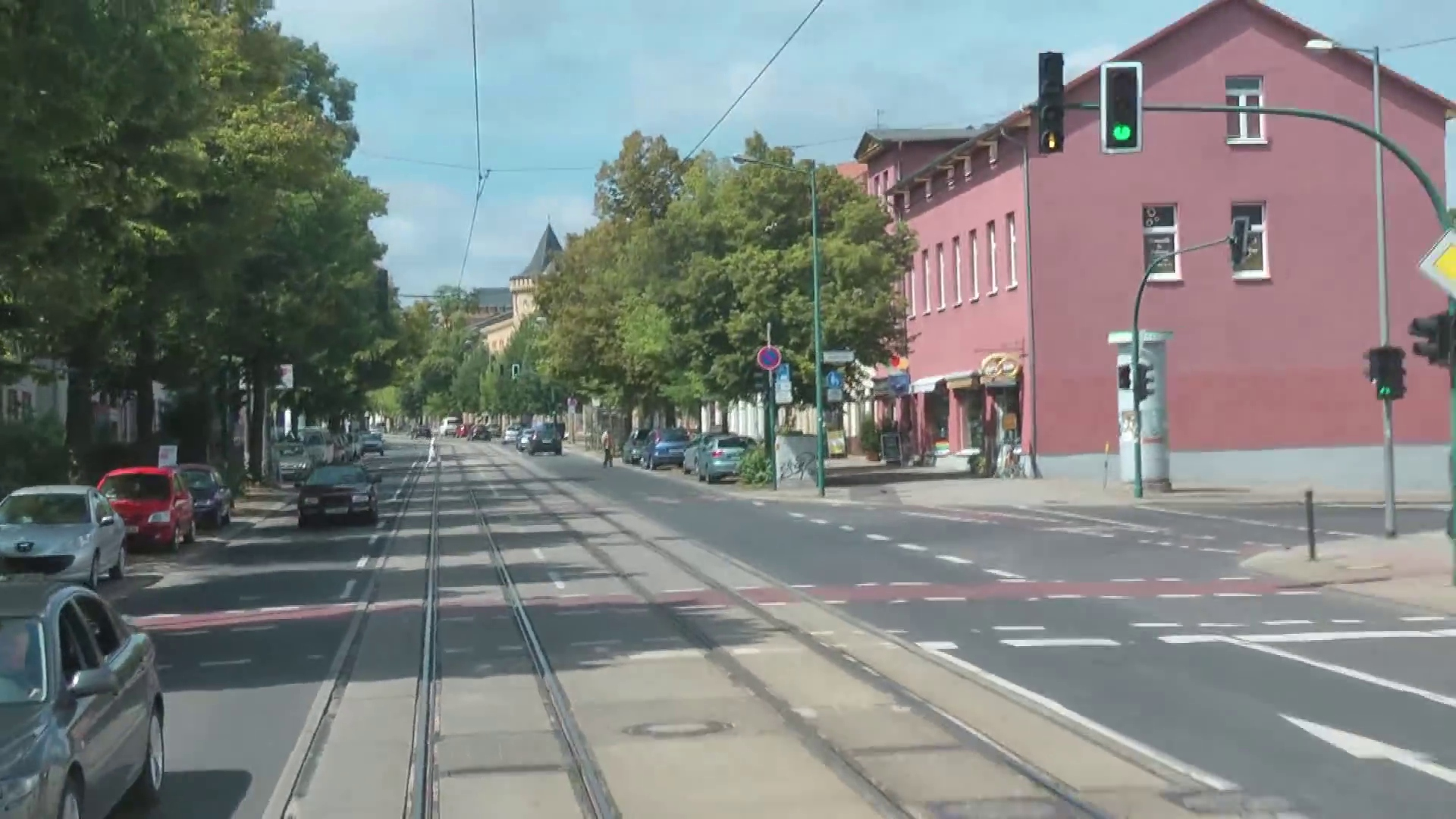}
        \caption{input image}
    \end{subfigure}
    \begin{subfigure}{0.15\textwidth}
        \includegraphics[width=\linewidth]{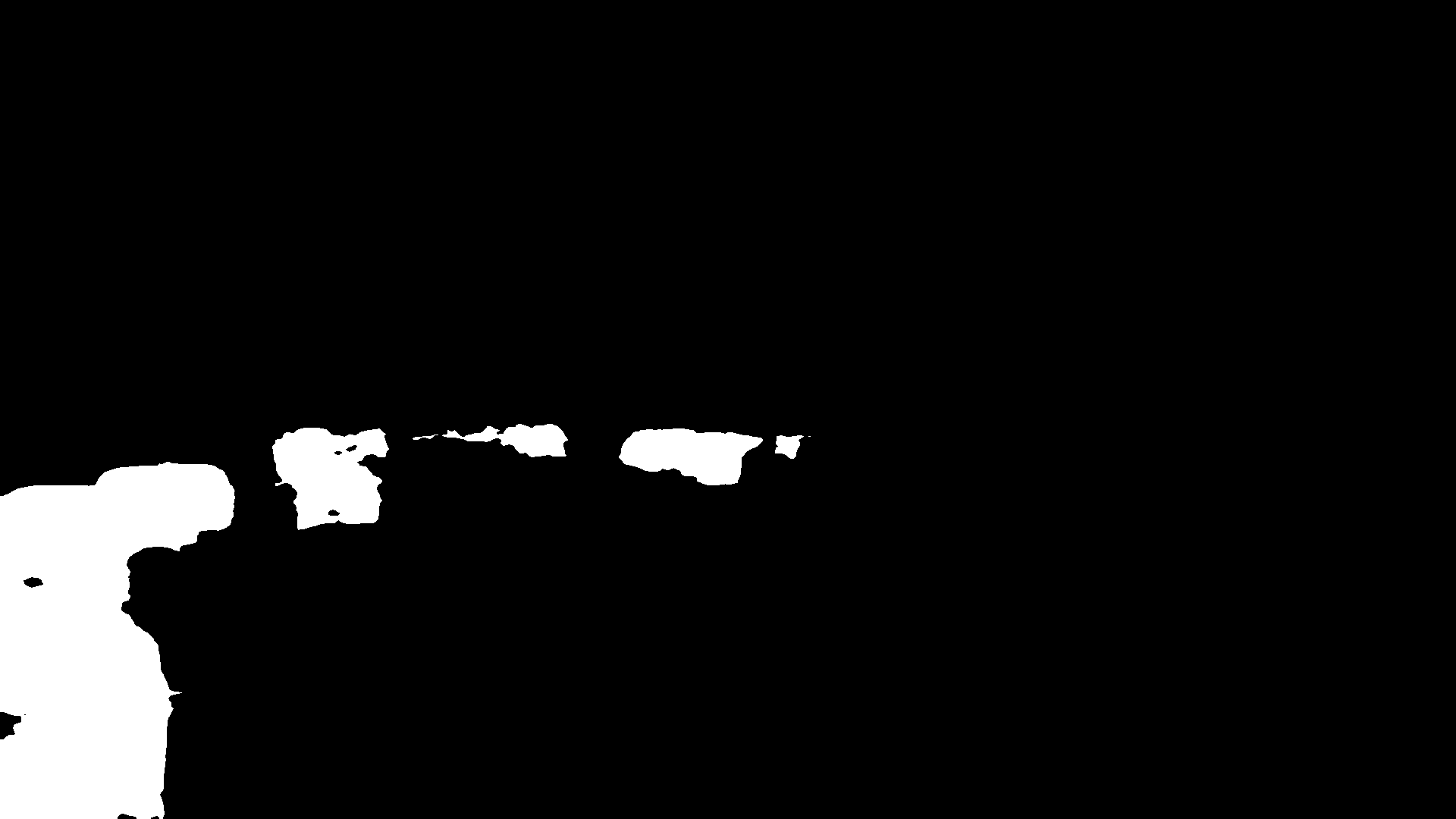}
        \caption{car}
    \end{subfigure}
    \begin{subfigure}{0.15\textwidth}
        \includegraphics[width=\linewidth]{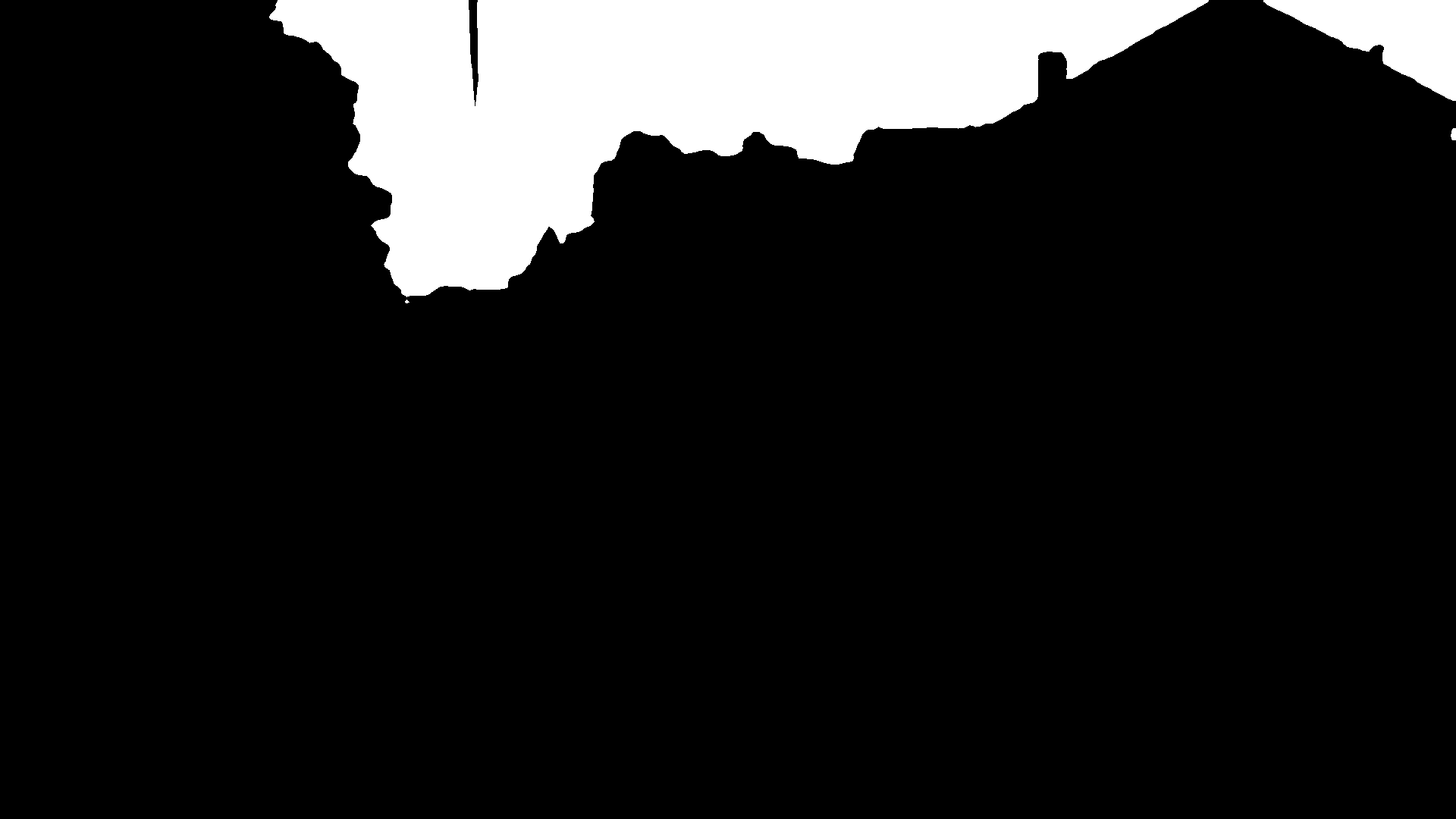}
        \caption{sky}
    \end{subfigure}

    \begin{subfigure}{0.15\textwidth}
        \includegraphics[width=\linewidth]{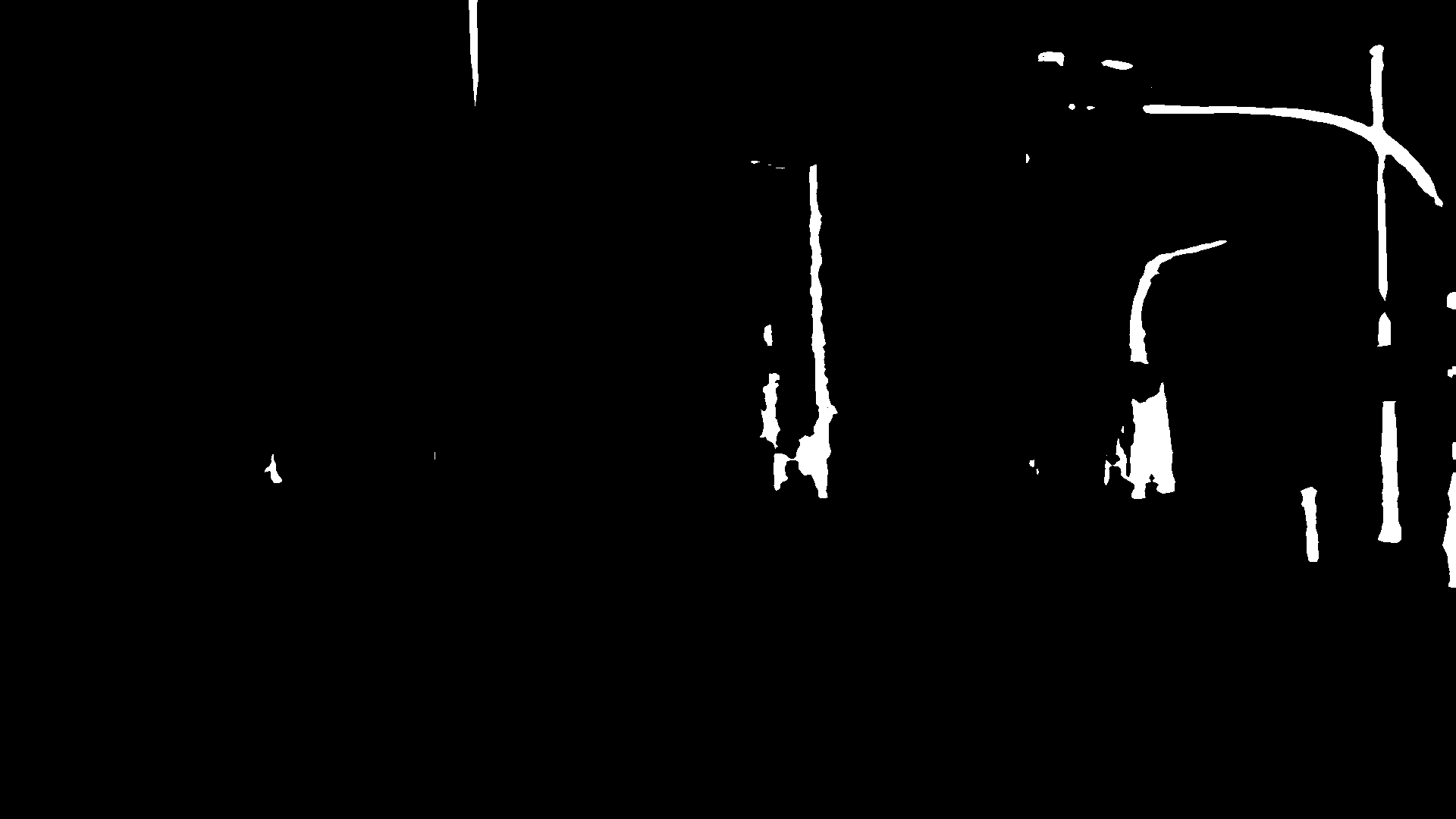}
        \caption{pole}
    \end{subfigure}
    \begin{subfigure}{0.15\textwidth}
        \includegraphics[width=\linewidth]{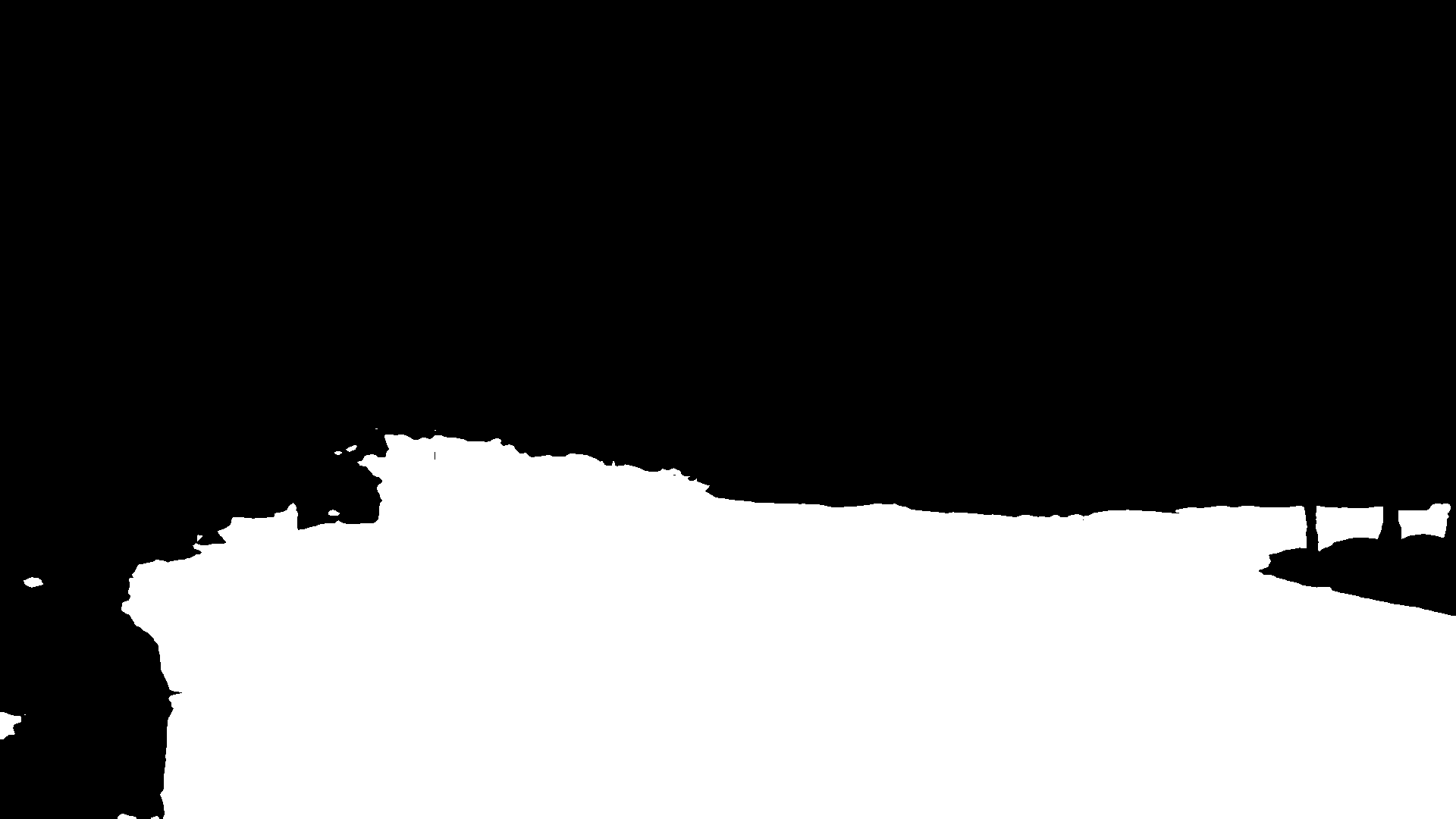}
        \caption{road}
    \end{subfigure}
    \begin{subfigure}{0.15\textwidth}
        \includegraphics[width=\linewidth]{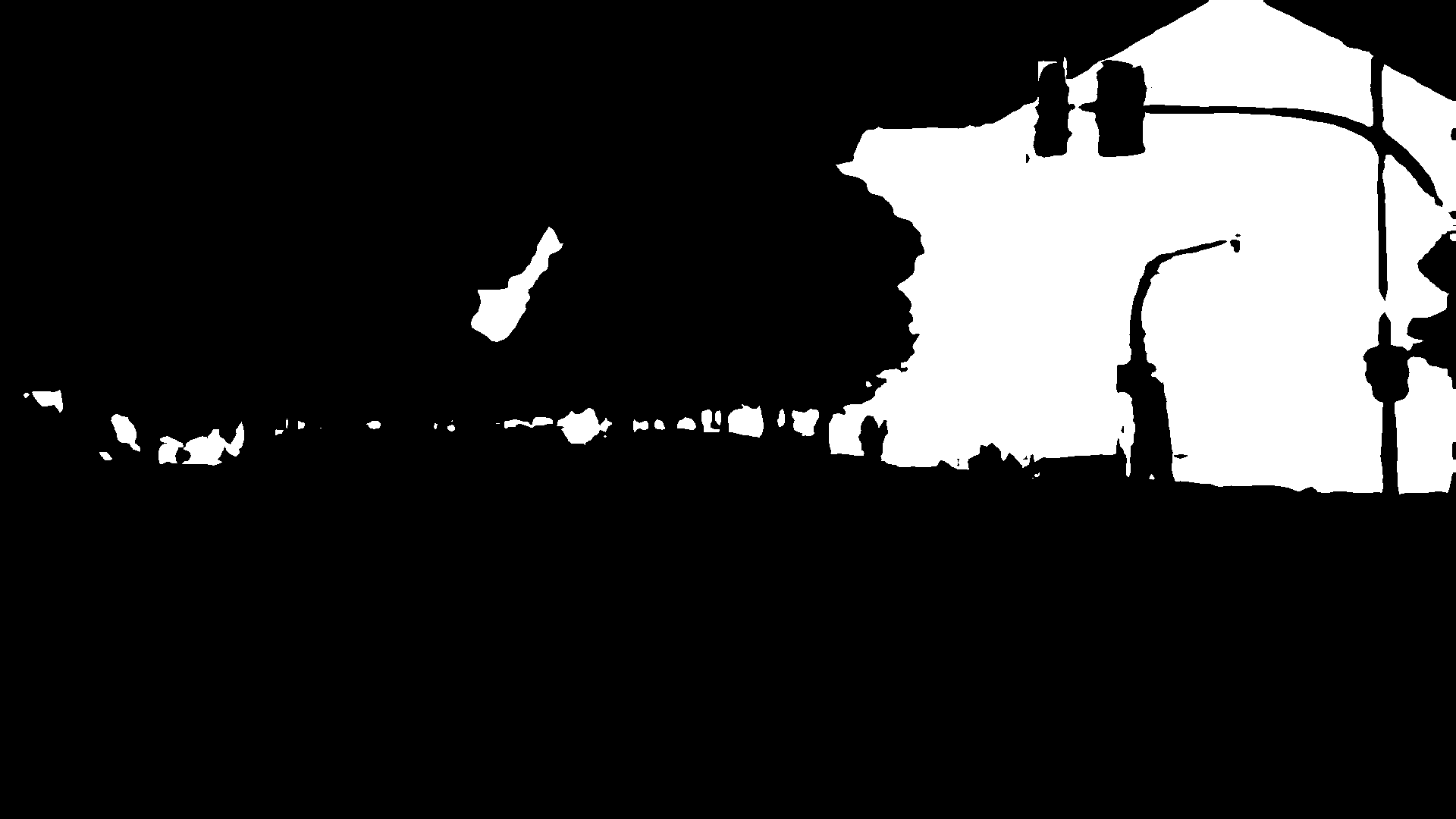}
        \caption{building}
    \end{subfigure}

    \begin{subfigure}{0.15\textwidth}
        \includegraphics[width=\linewidth]{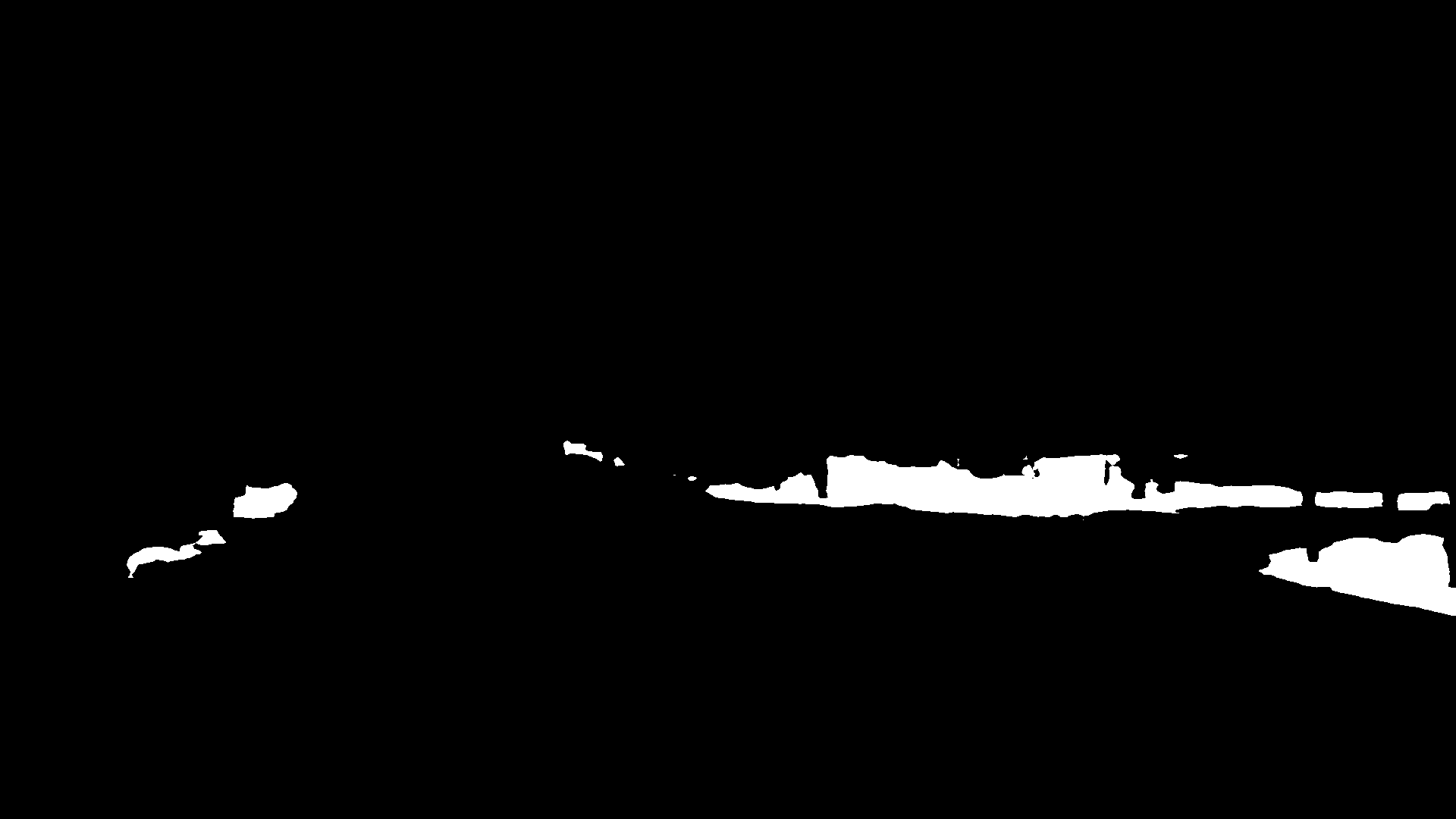}
        \caption{sidewalk}
    \end{subfigure}
    \begin{subfigure}{0.15\textwidth}
        \includegraphics[width=\linewidth]{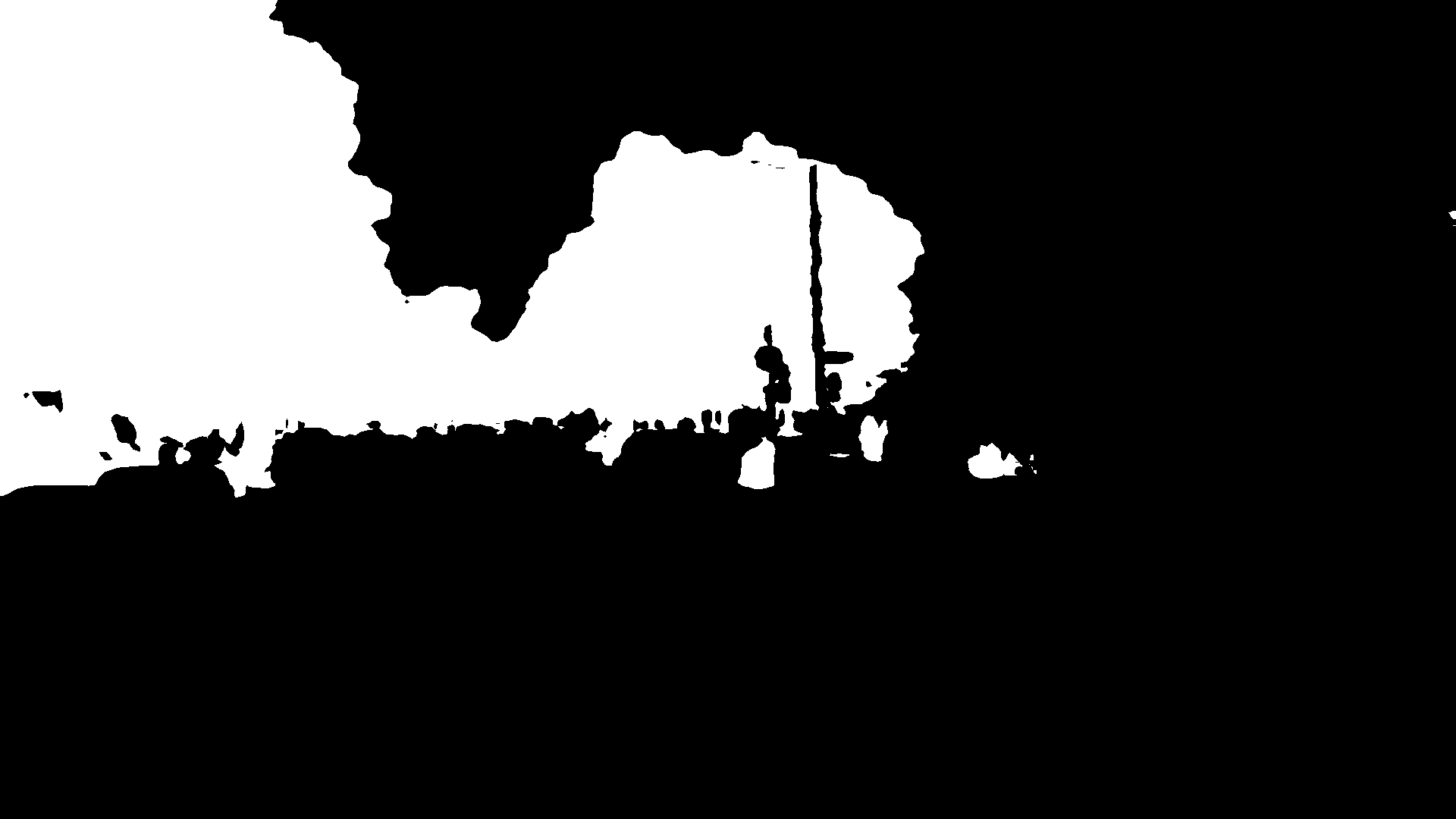}
        \caption{vegetation}
    \end{subfigure}
    \begin{subfigure}{0.15\textwidth}
        \includegraphics[width=\linewidth]{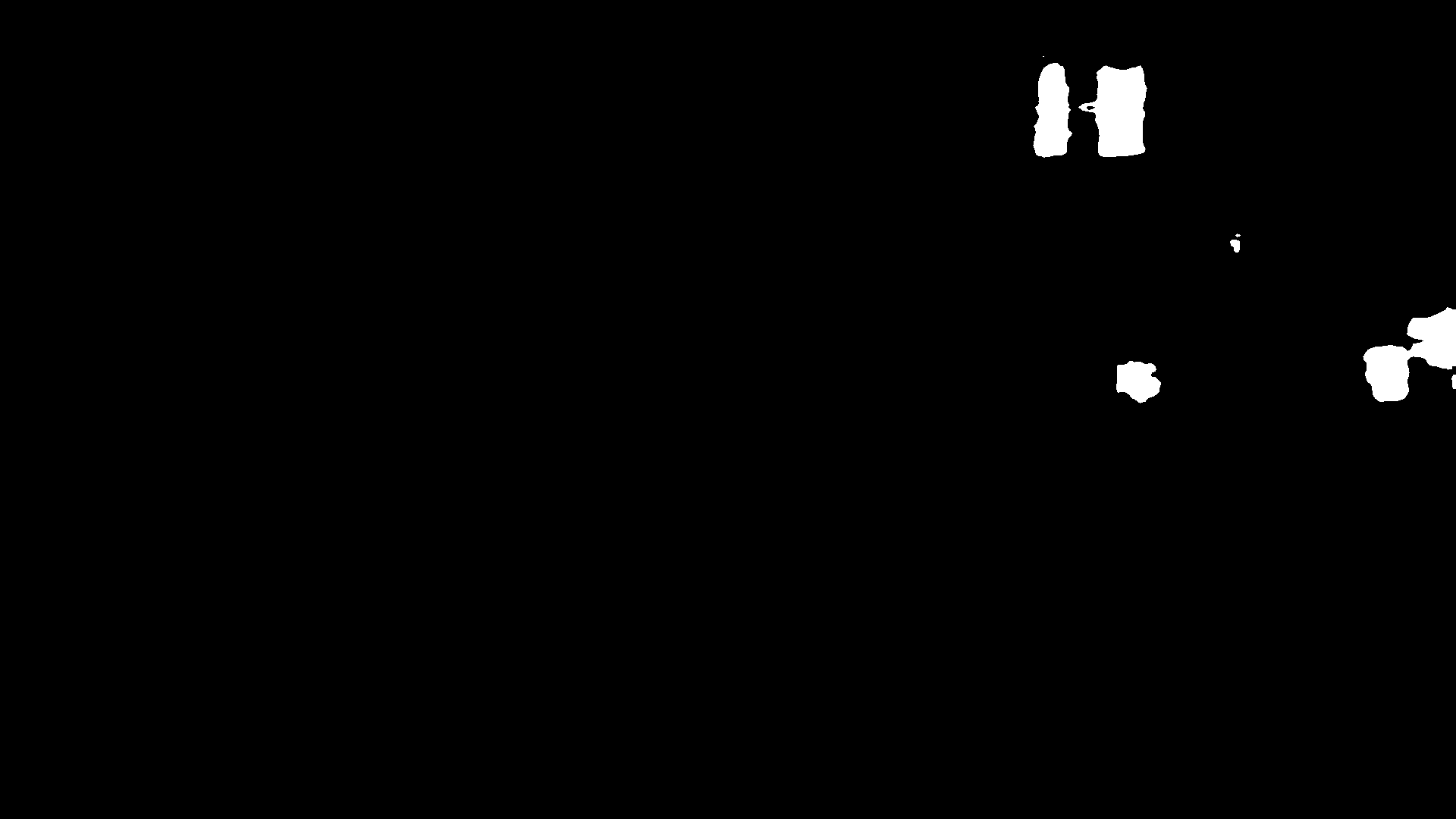}
        \caption{traffic lights}
    \end{subfigure}

    \caption{Given an input image (a), the segmentation model $f$ outputs binary masks, (b)–(i) each representing a distinct semantic class.}\label{fig:imgseg}
\end{figure}

Two types of constraint formulations can be used to define metamorphic relations (MRs) for a segmentation component.
The first is a strict equality constraint, which enforces pixel-level equivalence between the outputs $f(x)$ and $f(\tilde{x})$ under a transformation $g(x)$.
This yields $h\times w$ equality checks of the form: $M_\textit{eq} \coloneqq \delta(\theta, \tau)$ such that $\theta = f(x) - f(\tilde{x}), \tau=0$.
Such constraints can impose an overly rigid equivalence that may not hold in realistic conditions, particularly under mild perturbations or non-critical transformations.

To introduce tolerance, a relaxed IoU-based relation can be employed to measure the overlap between the predicted and actual segmentation outputs. This can be expressed as $M_\textit{IoU}(x, \tilde{x}; f) \coloneqq H(\theta, \tau)$, where $\theta = \frac{|B_i\land \tilde{B_i}|}{|B_i\lor \tilde{B_i}|}$.
Here, $B_i$ and $\tilde{B_i}$ are the $i$-th binary masks derived from $f(x)$ and $f(\tilde{x})$. 
This enforces the constraint $\theta \geq \tau$ and $\tau$ is the class-specific threshold defining the acceptable minimum intersection-over-union ratio.
A total of $k$ such relations, one per semantic class, jointly determines whether the segmentation output remains invariant under the transformation $g(x)$, thereby assessing the model’s robustness.


\subsubsection{Composite Metamorphic Relations.}
\label{sec:composite-metamorphic-relations}
For complex components, a single metamorphic property is often insufficient to characterize correct behaviour.
An object detector, for example, must not only detect the right class label for objects but also place bounding boxes accurately and with high confidence.
To capture such multifaceted requirements, we introduce \textit{Composite Metamorphic Relations (CMRs)}, which combine multiple individual metamorphic checks through logical conjunctions.
For each component $i$ in the system, let there be $n_i$ individual binary relations, $M_{i,j} : \mathcal{X} \times \mathcal{X} \to \{0, 1\}$, here, $j=1, \dots, n_i$ and $M_{i,j}=1$ if the $j$-th relation holds between the original input $x$ and its transformed variant $\tilde{x}$, and 0 otherwise.
These relations may capture diverse behavioural aspects, such as: \textbf{Output Integrity Checks} that ensure stability or invariance of output-level properties, e.g., Intersection-over-Union (IoU) between predicted and reference bounding boxes, consistency of confidence scores, or preservation of class labels and \textbf{Execution Trace Invariants} that ensure invariance of the system’s internal control or data flow.
For instance, a phantom call invariant asserts that the set of downstream modules invoked during execution remains unchanged under semantically equivalent transformations $g(x)$.
The individual relations are aggregated into a single composite score
$$S_i(x, \tilde{x}) = \prod^{n_i}_{j=1} M_{i,j} (x, \tilde{x})$$
Here, the product operator implements logical conjunction $S_i(x, \tilde{x})=1$ if and only if all constituent metamorphic relations hold, and $S_i(x, \tilde{x})=0$ otherwise.

This binary score thus provides an unambiguous, pseudo-oracle indicator of component-level correctness, allowing multiple relational properties to be verified simultaneously within a unified formal framework.
By construction, any violation in a constituent relation immediately flags a deviation in expected behaviour, offering both precision and interpretability in failure localization.
Additionally, user-defined specifications for each component can be encoded into relational properties, these individual checks are combined into a single composite module score holistically assesses the component's adherence to the user-defined requirements for each component throughout the system.
The score is 1 iff all individual metamorphic relations are satisfied, and 0 otherwise. The product therefore acts as the logical \texttt{AND} operator:
$$S_i(x, \tilde{x}) = \bigwedge^{n_i}_{j=1} M_{i,j}(x, \tilde{x})$$

\subsubsection{System-wide Composite Score.}
\label{sec:scoring-functions}
Given a system composed of $m$ components, each with its own composite score $S_i$, a global metamorphic consistency function is defined as $S(x, \tilde{x}) = \prod^{m}_{i=1} S_i(x, \tilde{x})$.
This aggregation captures the system’s overall adherence to metamorphic specifications under the transformation $g(x)$.
The score evaluates to 1 only if all constituent components satisfy their individual metamorphic relations, i.e., when the system as a whole behaves consistently under semantically invariant perturbations.
A zero-valued $S$ indicates at least one component has deviated from its expected relational behaviour.
The hierarchical composition of thus facilitates structured fault attribution: tracing system-level failures back to specific components and their violated metamorphic relations.
This structured aggregation enables scalable, interpretable assessment of correctness across multi-component AI systems without requiring access to an explicit ground-truth oracle.

\subsection{Statistical Fault Attribution}
\label{sec:fault-attribution}

Localizing the root cause of failures in compositional AI systems poses challenges analogous to debugging distributed microservice architectures, where system behaviour emerges from the interaction of independently executing components.
In such settings, \textit{distributed tracing} is employed to reconstruct end-to-end request paths, enabling the identification of the precise service or operation responsible for an observed fault.~\cite{xtrace,dapper_google}.
We adapt this principle to AI pipelines by instrumenting the execution flow represented as a computation graph composed of interdependent modules (e.g., perception, reasoning, decision-making) and tracing the sequence of activations captured within the graph as a tree corresponding to a single system run and its perturbed variants.

During execution, each module logs its activation, input-output behaviour, and any relational constraints defined by metamorphic relations (MRs).
These relations encode expected behavioural consistencies under controlled input transformations (e.g., invariance, monotonicity, equivalence).
By comparing observed outputs against MR specifications, the system detects deviations that indicate potential component-level anomalies within a larger compositional execution.

However, attributing fault to the first component that fails in a sequence may be misleading.
A downstream failure might be caused by a subtle, upstream error that did not violate a necessarily incomplete relational specification and propagate the erroneous state forward.
To mitigate this limitation, we introduce a statistical attribution framework that quantifies the contribution of each module to system-level inconsistencies.
This approach enables more principled fault attribution by modelling the causal influence of upstream deviations on downstream failures, thereby improving interpretability and diagnostic precision in complex AI systems.

To identify which components contribute most to failures (i.e., cases where $S=0$), we define a random variable $Z_i$ representing the metamorphic outcome of component $i$ over a distribution of inputs and transformations $Z_i=1-S_i(x, \tilde{x}), \quad \text{ where } (X, \tilde{X}) \sim \mathcal{D}$
Here, $S_i(x, \tilde{x})=1$ indicates that module $i$ satisfies its MR under transformation $\tilde{x}$, and $Z_i=1$ denotes a detected deviation.
This deviation is aggregated over a large number of test cases, conditioned on the occurrence of an end-to-end system failure.
Thus, the \textit{Failure Contribution} (FC) score for module $i$ is defined as the expected deviation of that module conditioned on a system-level failure $\text{FC}_i = \mathbb{E}_{(x, \tilde{x}) \sim \mathcal{D}}[Z_i\cdot\mathbb{I}(S(x, \tilde{x})=0)]
$
Empirically, over a dataset $\mathcal{D}$ and a set of perturbations $\mathcal{P}$, the FC score is estimated as: $$\text{FC}_i = \frac{\sum_{x\in\mathcal{D}}\sum_{g\in\mathcal{P}}Z_i\cdot\mathbb{I}(S(x, g(x)))}{\sum_{x\in\mathcal{D}}\sum_{g\in\mathcal{P}}\mathbb{I}(S(x, g(x)))}$$
This measures how frequently component $i$ violates its specification given that the system fails, effectively capturing its conditional responsibility for test inputs sampled from a dataset $d \in \mathcal{D}$ and perturbations sampled from a set $g\in \mathcal{P}$.
This metric augments simple temporal precedence to include statistical correlations measuring how strongly a component's failure is associated with the system's failure.

Further, an interpretable blame assignment is performed by normalizing these scores across all $m$ modules in the system to derive interpretable \textit{attribution weights}: $\alpha_i = \frac{\text{FC}_i}{\sum_{j=1}^m \text{FC}_j}$.
The normalized $\alpha_i$ quantifies each component’s relative contribution to system-level unreliability.

In practice, this framework simulates sequential fault attribution along the computation trace, ordered by temporal precedence but augmented with statistical evidence.
When the metamorphic specifications are sufficiently sound and complete, this trace-based attribution becomes analogous to causal fault attribution, allowing developers to prioritize debugging on the components most likely to be the true root cause rather than merely the point of failure propagation.
This formulation bridges correlation-based and causal reasoning by grounding component attribution in both behavioural constraints and statistical dependencies.
The robustness of such a system is checked against a collection of such properties and the subset of models for which the properties are adversarial are isolated.

To illustrate this, consider the Vision System developed for Autonomous Railway Maintenance Vehicles~\cite{albinmasters}.
This system processes real-time video streams from an ego-perspective camera and consists of two key modules:
(1) a \textit{railway track localization module}, and
(2) an \textit{object detection module} responsible for identifying obstacles near the track.
The outputs of the detection module trigger downstream classifiers to generate prognostic reports and control signals for navigation.
When evaluated under controlled perturbations (e.g., lighting variations, occlusions), our framework isolates cases where failures in track localization indirectly cause misdetections in the object detection pipeline, highlighting causal fault propagation that would otherwise be obscured by purely temporal analysis.

\section{State-based Execution Trace Analysis}
\label{sec:seta}

Any compound AI system can be modelled as a state-transition system, wherein each state is roughly equivalent to a component of the system that receives some data, performs some computation and routes the resultant output to one or many states.
Let an AI system $\mathcal{S}$ be defined by a tuple:
$$
\mathcal{S} = (Q, I, \Phi, \mathcal{M}, \mathcal{R}, \mathbf{S})
$$
where $Q = \{q_1, q_2, \dots, q_n\}$ is a finite set of \textbf{states}, each representing a computational module; each state $q_i$ representing a computational unit within $\mathcal{S}$. A state is represented as a tuple: $q_i = (f_{i}, R_{i}, S_{i})$, where $f_{i}$, $R_{i}$ and $S_{i}$ are the Model, Routing function and Scoring functions respectively.
$I \in Q$ is the \textbf{initial state}, which receives external inputs.
$\Phi \in Q$ is the set of \textbf{ terminal states}, producing final outputs.
$\mathcal{F} = \{f_q : \mathbb{R}^d \to \mathbb{R}^{d'} \mid q \in Q\}$ is the set of \textbf{model functions}, one per state, each defining a module’s computation for that state.
$\mathcal{R} = \{R_q : (Q \times\mathbb{R}^{d'}) \to 2^{(Q \times\mathbb{R}^{d'})} \mid q \in Q\}$ is the set of \textbf{routing functions}, determining which successor modules are activated given a state and its output and the input data to be routed to them.
$\mathbf{S} = \{S_q : (\mathbb{R}^{d'} \times \mathbb{R}^{d'}) \to \{0, 1\} \mid q \in Q\}$ is the set of \textbf{scoring functions} comparing reference and perturbed outputs for metamorphic consistency.
Given input $x \in \mathbb{R}^d$, the system’s runtime behaviour can be captured as an \textit{Execution Trace Tree} $T(x)= (V, E)$, where each node $v = (q, x_v, y_v, S_q) \in V$ corresponds to: (i) the module $q \in Q$ activated during execution, (ii) the module input $x_v$, (iii) its output $y_v = f_q(x_v)$ and, (iv) the scoring function $S_q(y_v, \tilde{y}_v)$ comparing reference and perturbed runs.
Edges $(v, w) \in E$ are induced by the routing functions $R_q$, with the tree rooted at the initial input node $I$.
The same process is repeated for perturbed input $\tilde{x}$ to obtain $T(\tilde{x})$, which is then aligned against $T(x)$ for relational evaluation.

A model is considered robust to perturbation $\tilde{x}$ if, for every node $v$ in $T(x)$ that is also activated in $T(\tilde{x})$, $S_q(y_v, \tilde{y_v}) = 1$ and the corresponding routing decisions remain invariant.
Deviations where $S_q(y_v, \tilde{y}_v) = 0$ or the activation set diverges (i.e., a module was expected but not triggered, or vice versa) are recorded as component-level faults.

Considering parent and child states $q_i$ and $q_j$, their respective states contain models $f_i : \mathbb{R}^{d_i} \to \mathbb{R}^{d_i'}$ and $f_j : \mathbb{R}^{d_j} \to \mathbb{R}^{d_j'}$.
If $q_j \in R_i(f_i(x))$, i.e., $q_i$ routes to $q_j$, then $d_i' = d_j$ or the output-input dimensionality of the models should be compatible.

To instantiate the proposed statistical attribution framework, the \textit{State-based Execution Trace Analysis (SETA)} procedure is implemented.
Given a test dataset $\mathcal{D}$ and a set of perturbations $\mathcal{P}$, the framework executes the system $\mathcal{S}$ over each input–perturbation pair $(x, g(x))$, dynamically unrolling the corresponding computation tree $T(x)$.
For every input, the framework records which modules were activated, their expected activation according to routing functions, and the corresponding metamorphic deviations $S_i(x, \tilde{x})$.
By aggregating these deviations over all runs, SETA estimates a failure count score for each module and normalizes it into an attribution coefficient $\alpha_i$, capturing the empirical contribution of module $i$ to system-level failures.
Algorithm~\ref{alg:seta} outlines this computation.

\begin{algorithm}
\caption{SETA Fault Attribution}
\label{alg:seta}

\textbf{Input:} System $\mathcal{S}$ with modules $f_1, \dots, f_n$, Test Dataset $\mathcal{D}$ \\
\textbf{Parameter}: Perturbation Set $\mathcal{P}$, Scoring Functions $\mathcal{S}$\\
\textbf{Output:} {Attribution scores} $\alpha_1, \dots, \alpha_n$
\begin{algorithmic}[1]
\STATE Initialize $\texttt{FC}[i] \gets 0$ for all $i \in [1,n]$
\STATE \texttt{TotalFailures} $\gets 0$
\FOR{\textbf{each} $x \in \mathcal{D}$}
\STATE Record reference execution trace $T(x)$ and activated modules $A(x)$
\FOR{\textbf{each} perturbation $g \in \mathcal{P}$}
\STATE $\tilde{x} \gets g(x)$
\STATE Record perturbed trace $T(\tilde{x})$ and activated modules $A(\tilde{x})$
\STATE Compute system-level score $S(x,\tilde{x}) = \prod_i S_i(x,\tilde{x})$
\IF{$S(x,\tilde{x}) = 0$}
\STATE $\texttt{TotalFailures} \gets \texttt{TotalFailures} + 1$
\FOR{$i \gets 1$ \textbf{to} $n$}
\STATE $\texttt{activatedRef} \gets [f_i \in A(x)]$
\STATE $\texttt{activatedPert} \gets [f_i \in A(\tilde{x})]$
\STATE $\texttt{deviation} \gets [S_i(x,\tilde{x}) = 0]$
\IF{$\texttt{activatedRef}$ \textbf{or} $\texttt{activatedPert}$}
\STATE $\texttt{FC}[i] \gets \texttt{FC}[i] + (\texttt{deviation} + (\texttt{activatedRef} \oplus \texttt{activatedPert}))$
\ENDIF
\ENDFOR
\ENDIF
\ENDFOR
\ENDFOR
\STATE Normalize $\texttt{FC}[i] \gets \frac{\texttt{FC}[i]}{\texttt{TotalFailures}}$ for all $i$
\STATE Compute $\texttt{TotalFC} \gets \sum_j \texttt{FC}[j]$
\FOR{$i \gets 1$ \textbf{to} $n$}
\STATE $\alpha_i \gets \frac{\texttt{FC}[i]}{\texttt{TotalFC}}$
\ENDFOR
\STATE \textbf{return}~$ \alpha_1, \dots, \alpha_n$
\end{algorithmic}
\end{algorithm}
\noindent While the SETA framework provides a structured and interpretable approach to fault attribution, several limitations remain.
Metamorphic relations may be incomplete or non-causal, leading to false negatives when violations occur outside the defined property space.
Deviations estimated from limited perturbation sets or biased datasets may misrepresent component influence.
Moreover, since execution traces capture observed rather than true causal dependencies, attribution accuracy depends on the fidelity of recorded routing and state transitions.
Finally, the statistical aggregation assumes independence between perturbation effects, which may not hold in tightly coupled systems.
These challenges motivate future extensions involving causal trace reconstruction, adaptive property learning, and uncertainty-aware attribution.

\section{Experimental Setup}

The SETA framework is implemented as an instrumentation layer integrated into the system runtime.
During execution, each module registers a lightweight tracing hook that records the input and output tensors, metadata such as confidence scores and class logits, dynamic routing decisions determined by $R_q$ and the identity of downstream modules invoked.
This enables construction of the execution trace tree $T(x)$ without modifying the underlying model architectures.

Perturbations are drawn from a configurable set $\mathcal{P}$ that can contain both \textit{input-space transformations} (e.g., Gaussian noise, brightness scaling, or occlusion for vision models) and \textit{intermediate-state perturbations} (e.g., feature masking or latent-space jitter).
For each perturbation $\tilde{x} = g(x)$, the framework replays the full computation under identical routing conditions, recording the perturbed execution $T(\tilde{x})$.
All runs are executed deterministically to ensure reproducibility of traces and facilitate node-level alignment.

After execution, the framework aligns corresponding nodes between $T(x)$ and $T(\tilde{x})$ using structural and module identifiers, then evaluates each module’s composite MR score $S_i(x,\tilde{x})$.
Failures and routing inconsistencies are aggregated using Algorithm~\ref{alg:seta} to produce the normalized attribution vector $\boldsymbol{\alpha} = [\alpha_1, \dots, \alpha_n]$.
All results, including trace trees, MR evaluations, and failure logs, are persisted in a structured event store for downstream analysis and visualization.

To evaluate the functioning of the proposed framework, two systems that implement computer vision tasks have been considered: a Railway Vision System~\cite{albinmasters}, a simple ensemble model and an OCR model. We describe the vision system application in detail, other examples can be found in the \href{https://github.com/mojibake00/vision_system}{online repository}.

\subsection{Vision System}
\label{sec:vision-system}
The first system considered is the Vision System~\cite{albinmasters} as outlined in Section~\ref{sec:background}.
The system contains 6 modules: 1 object detector and 5 classifier models.
The object detector is a YOLOv4-tiny~\cite{yolov4tiny}, trained to detect objects specific to the railway context, such as traffic signals, road crossing, catenary poles, etc., along with some standard objects.
There are also a few special classes ({speed\_sign}, {main\_signal}, {road\_crossing\_signal}, etc.) that invoke one of the 5 remaining classification models on being predicted, in order to provide additional details regarding the object.
For e.g., when a {speed\_sign} object is detected, the classification model corresponding to the label is invoked in order to determine the speed limit mentioned in the sign.
All models in the system are lightweight and modular, being preferable to a single large model when deployed on power-constrained devices with limited compute.
The proposed framework can determine the number of errors and attribute them to their respective models.

\begin{figure*}
    \centering
    \includegraphics[width=0.5\linewidth]{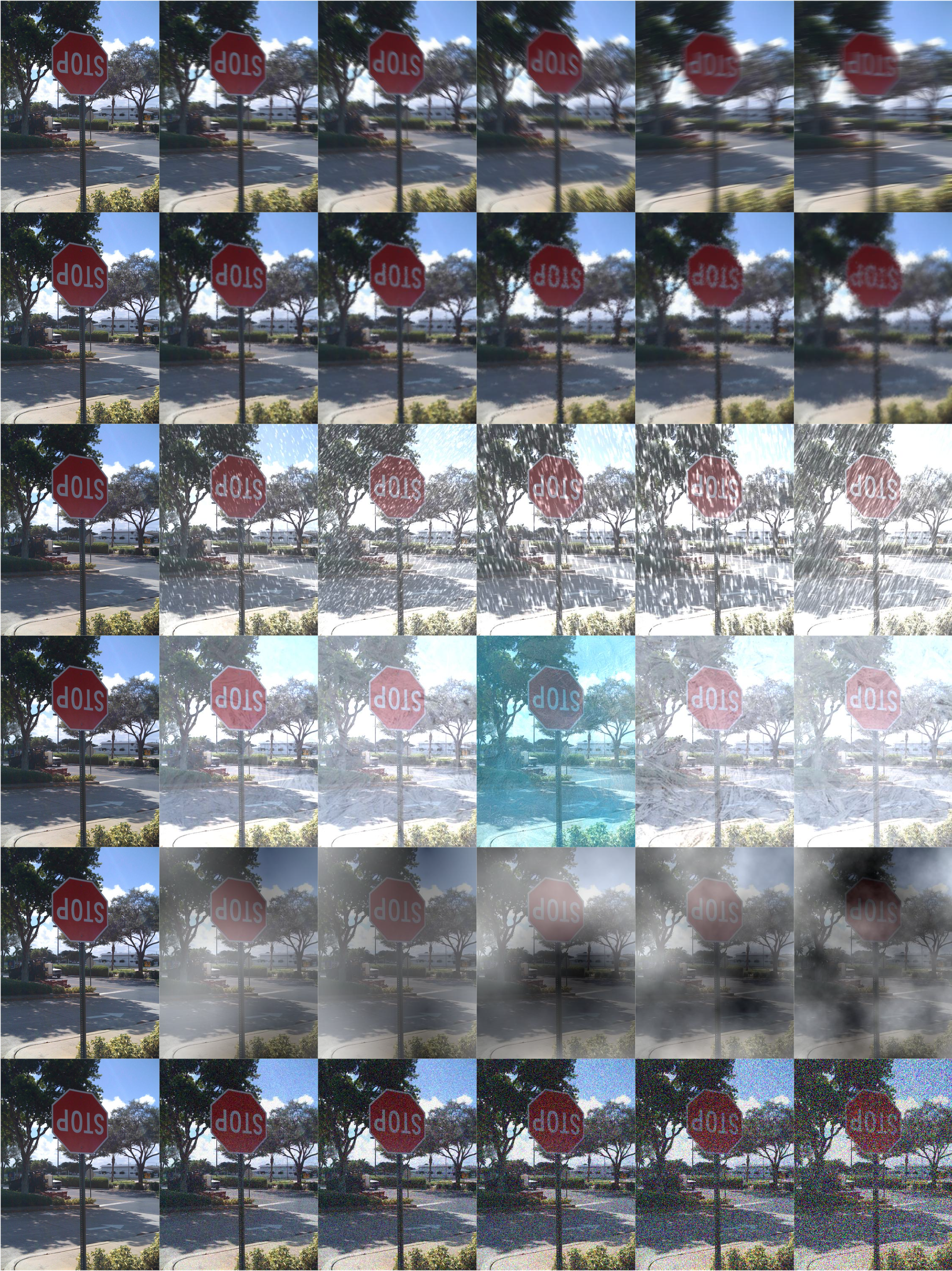}
    \caption{A variety of weather and noise based (\texttt{zoom\_blur}, \texttt{glass\_blur}, \texttt{snow}, \texttt{frost}, \texttt{fog} and \texttt{gaussian\_noise}) perturbations have been utilized to generate an extended synthetic test dataset that tests the autonomous railway system for robustness against weather.}
    \label{fig:imagecorruptions}
\end{figure*}

\subsubsection{Test Dataset Generation}
\label{sec:data}

To facilitate testing in oracle-less scenarios, synthetic test cases corresponding to actual test cases are generated.
These examples are generated using the \textit{imagecorruptions}~\cite{michaelis_imagecorruptions_2020,hendrycks2019robustness} library that enables simulations of different weather conditions and image noise varieties.
This set of tools have been specifically chosen in the context of testing autonomous driving systems that must be robust to different weather phenomena.
The \textit{imagecorruptions} library contains a range of perturbations that can be applied at different scales to the input image.
The predictions from the unaltered input image is stored as a reference input.
Individual component predictions corresponding to the perturbed inputs (at various scales) are then compared with the reference in order to ascertain whether the output is within a tolerable threshold of deviation from the reference output.

\subsubsection{Metamorphic Relations for Vision System}
\label{sec:MRvis}

\begin{figure}[h]
    \centering
    \includegraphics[width=0.3\linewidth]{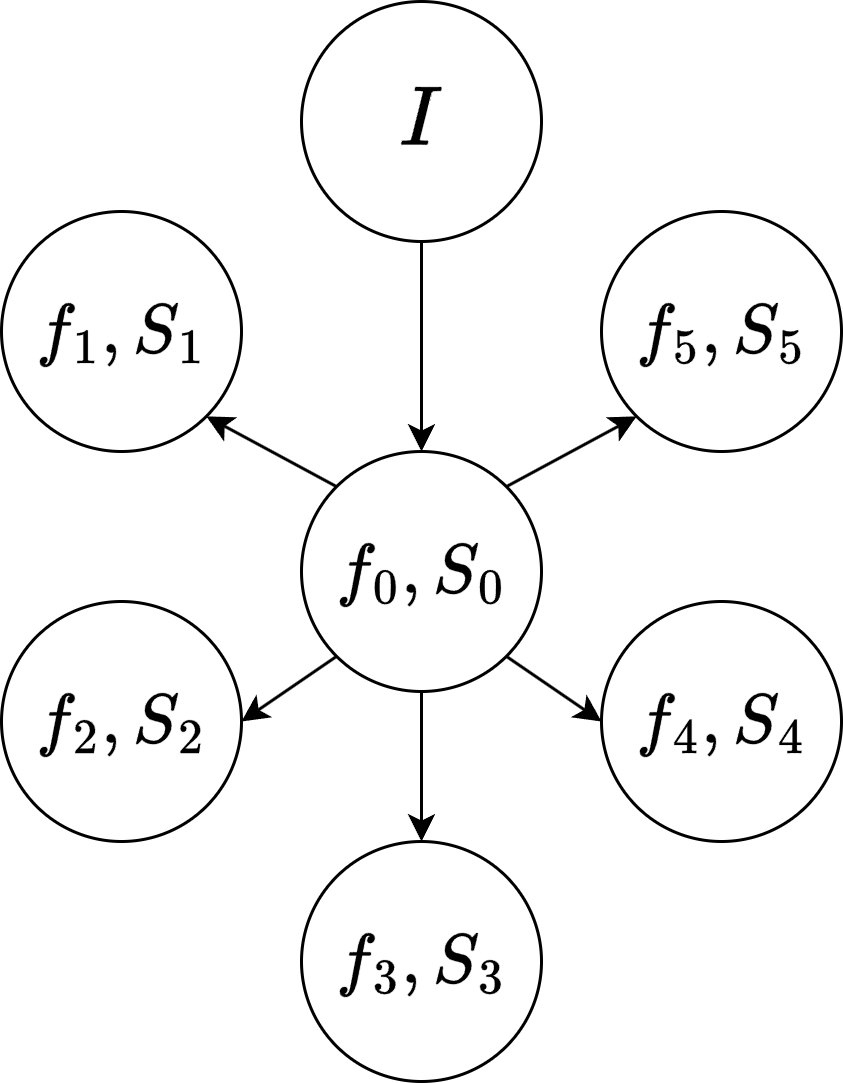}
    \caption{DAG representing the topology of the Vision System}
    \label{fig:topo}
\end{figure}

The Vision System has multiple vision oriented modules in the topology as illustrated in Figure~\ref{fig:topo}.
As described in the Background section, each individual component has its own set of Metamorphic Relations (MRs) that need to be satisfied in order for the output of the component to be considered consistent with the reference input.
\noindent Each component is paired with a scoring function, which are metamorphic relations~\cite{compositeMR} composited together through logical conjunctions.
These functions check for key properties that should remain invariant for each model.
For the illustrative examples provided, there are two classes of DNN models utilized.
A brief description of them can be found below:
\begin{itemize}
    \item \textbf{Object Detection} $f_0$ is an object detector, specifically a modified YOLOv4-tiny~\cite{yolov4tiny}, that is capable of detecting standard everyday objects alongside objects specific to the use case of autonomous railway vehicles.
    \item \textbf{Image Classifiers} models $f_1 \dots f_5$ are simple image classifier models that receive inputs from $f_0$ and provide further signals relevant to controlling aspects of the autonomous vehicle.
\end{itemize}
Each scoring function $S_i(x, \Tilde{x})$ has the generic form $\prod^{n_i}_{j=1} M_{i,j}$, where each $M_{i,j}$ refers to a property that will be checked per test instance.
For classes of models that perform the same task, or provide similar form of outputs, the MRs will be comparable with the only changes being any parameters that are required to check said properties.

\subsubsection{Scoring function for $f_0$}
The properties that must be satisfied for an object detection model are as follows:
\begin{itemize}
    \item $M_{0,1}$ : The set of detections (\textbf{D}) in the perturbed image must be a subset of the original detections (\textbf{O}).
    \item $M_{0,2}$ : Persisting detections ($\textbf{D}\subseteq\textbf{O}$) should have \textit{Intersection over Union (IoU)} values $> 0.9$.
    \item $M_{0,3}$ : For the subset of persisting detections, the class labels must remain unchanged.
    \item $M_{0,4}$ : The set of invoked downstream classifiers must be a subset of those invoked by the source image. This is included in the previous condition as the class labels refer to the modules downstream.
\end{itemize}
Another additional constraint that we can set on top of these already existing metamorphic relations is the difference between the predicted and reference confidence of the model.
In scenarios where the confidence of a model is available to the tester, it can be utilized as a constraint that can be used to ensure that the confidence of the model does not deviate too much for small perturbation in the input.

\subsubsection{Scoring function for $f_1 \dots f_5$}
The properties that must be satisfied for the image classifier models are as follows:
\begin{itemize}
    \item $M_{i,1}$ : All image classifiers have a single similar property that they must always satisfy. The labels of the outputs must always match the reference label. This is necessary and sufficient for checking whether the output of an image classifier is correct or not.
\end{itemize}
Some additional metamorphic relations that can checked for classifier models is related to the softmax probabilities or logits of the models, however, in a lot of inference scenarios these details may not necessarily be available for testing the integrity of the output.

\noindent
\paragraph{\textbf{Datasets.}}
The perturbation techniques chosen for this work are collectively called \textit{imagecorruptions}~\cite{michaelis_imagecorruptions_2020}, and are largely adapted to test object detector models.
These techniques extend classical corruption techniques for classification dataset~\cite{hendrycks2019robustness} that are used to generate augmented test datasets.
The techniques simulate different weather conditions under which autonomous driving systems must operate safely, thus being highly relevant to testing robustness of the Vision System.
There are 15 different corruption techniques provided by \texttt{imagecorruptios}, namely, \texttt{gaussian\_noise}, \texttt{snow}, \texttt{frost}, \texttt{fog}, \texttt{motion\_blur}, \texttt{glass\_blur}, etc., each having 5 severity levels.

The ensemble model is tested on a corrupted version of the CIFAR10 test data. For the Vision System, no proper pre-existing datasets exist that specifically address our use case. Hence, by sampling data from the RailSem19~\cite{railsem19} dataset that depicts the railway scene imagery from the ego-perspective of a train, an augmented dataset is generated to act as a test benchmark for the Vision System. 

We evaluate SETA on two representative multi‑DNN vision pipelines: (i) an Ensemble Classification Model on CIFAR‑10, and (ii) an Autonomous Rail‑Inspection Vision System.
The framework is utilized to compute predicted accuracies of each of the models and compare them with actual accuracies to determine the efficacy of the framework.
Since the framework is architecture agnostic, every constituent model of the system under analysis is treated as a black-box, with only the input and output formats available.
\begin{figure*}[htbp]
\begin{subfigure}{.33\textwidth}
  \centering
  \includegraphics[width=\linewidth]{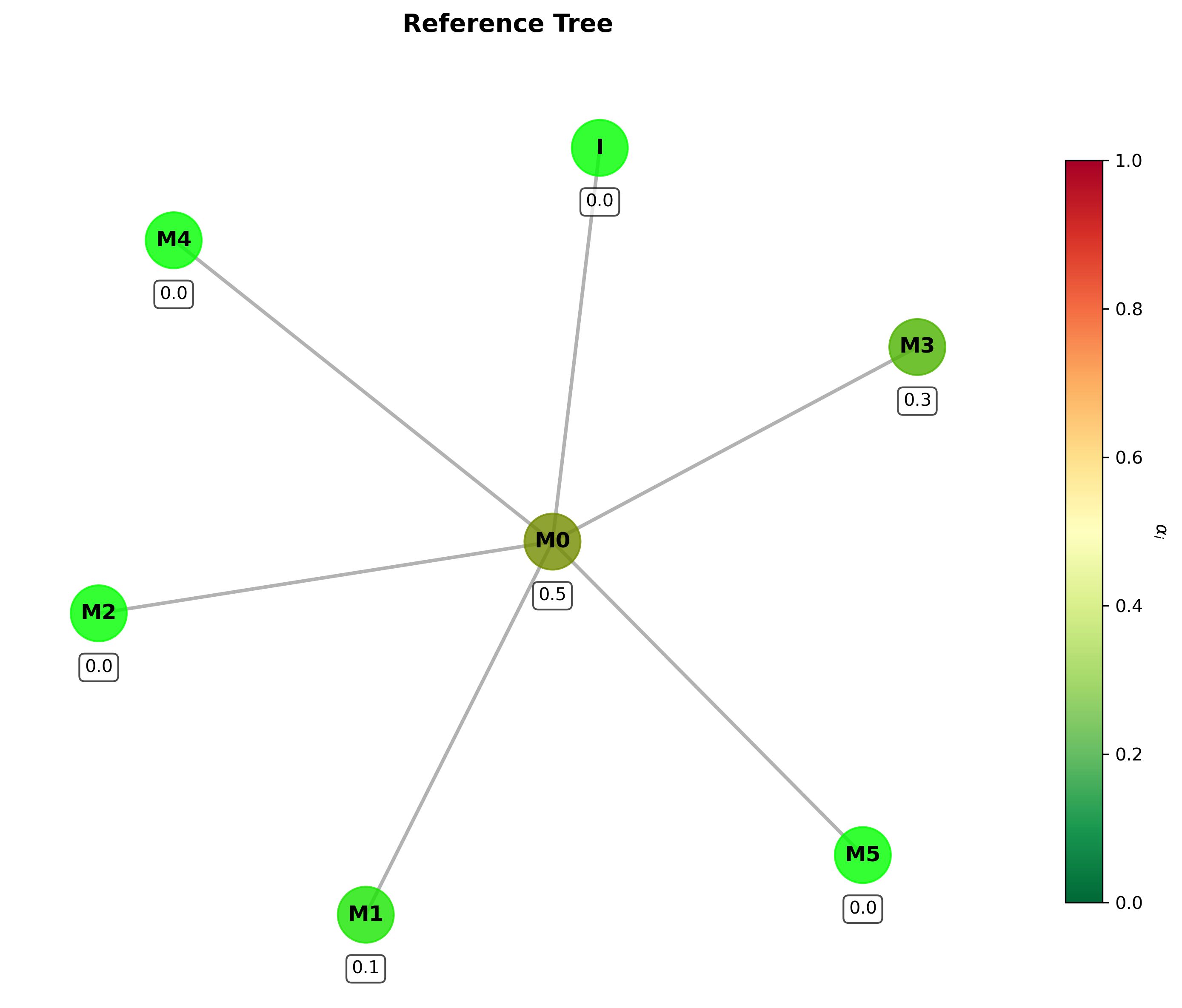}
  \caption{\texttt{fog}}
  \label{fig:tree1}
\end{subfigure}%
\begin{subfigure}{.33\textwidth}
  \centering
  \includegraphics[width=\linewidth]{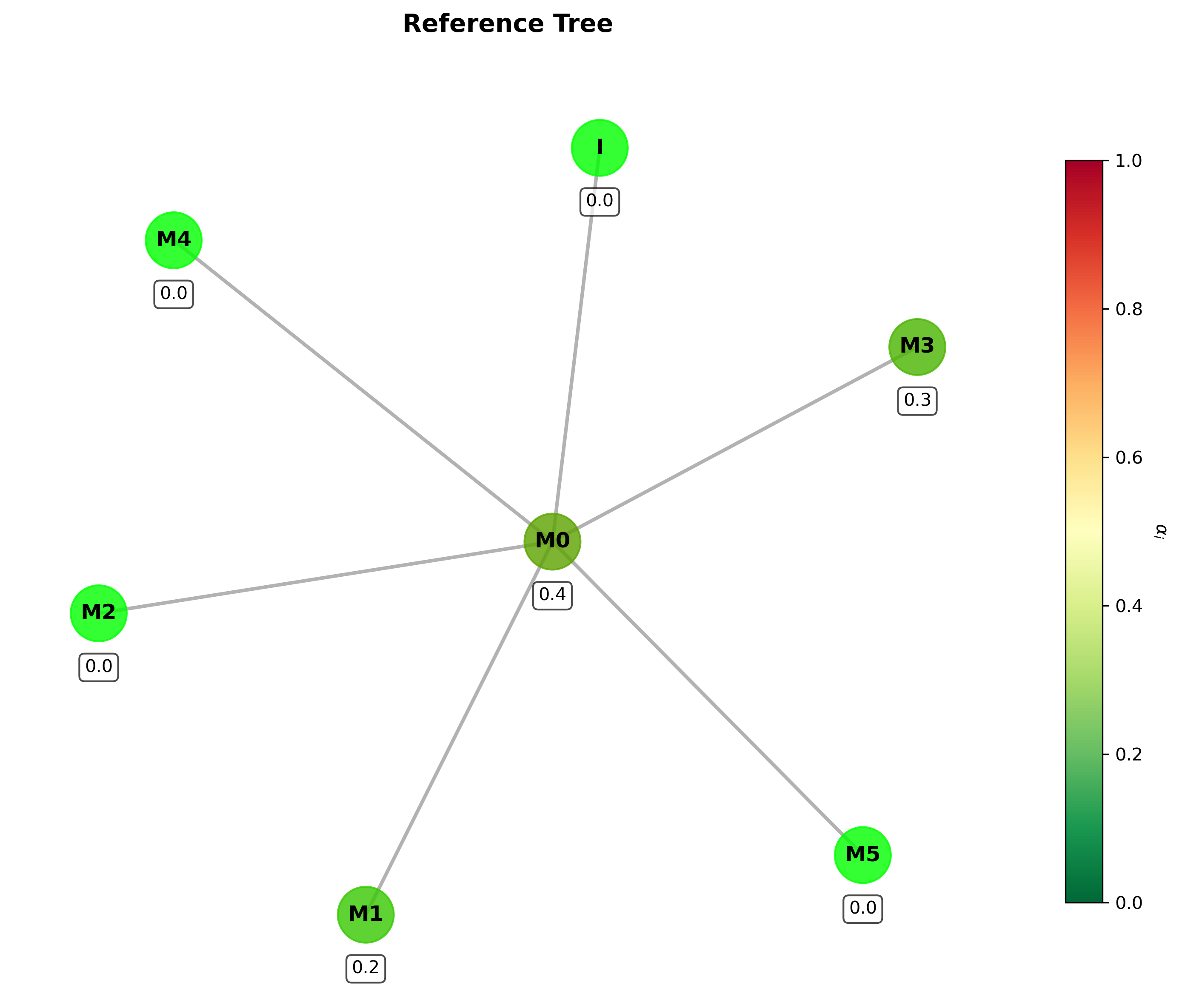}
  \caption{\texttt{glass\_blur}}
  \label{fig:tree2}
\end{subfigure}
\begin{subfigure}{.33\textwidth}
  \centering
  \includegraphics[width=\linewidth]{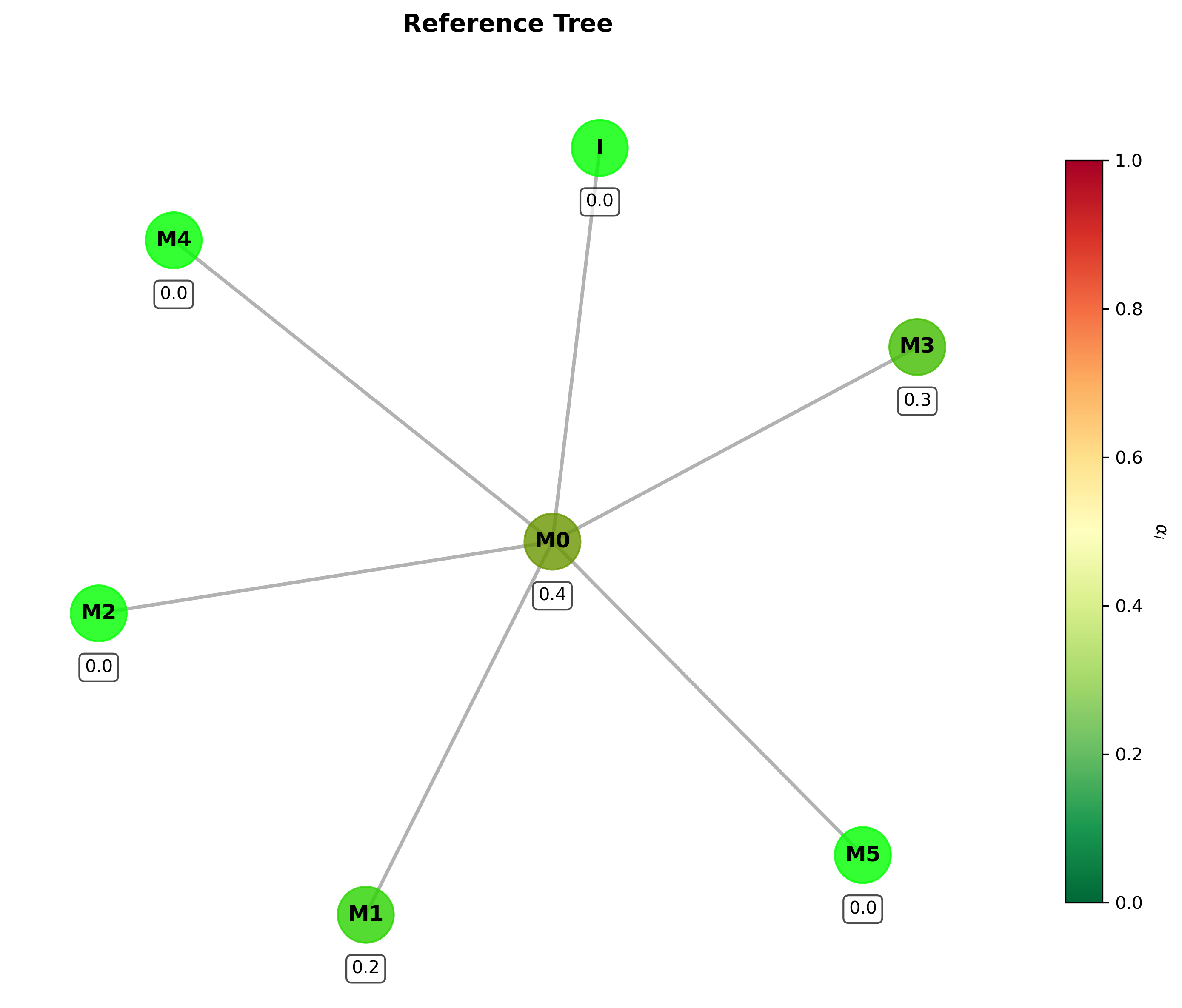}
  \caption{\texttt{motion\_blur}}
  \label{fig:tree3}
\end{subfigure}%
\caption{Response of the Vision System to various image corruption techniques.}
\label{fig:tree}
\end{figure*}
\noindent For a given CIFAR10 test dataset augmented by image corruptions, the framework is able to determine the approximate accuracy of each model by keeping track of every non-robust output against the total number of invocations for the model.
As can be seen from Table~\ref{table:ensemble}, the predicted values are within the range of 2-3\% and is used to pick the worst performing model, which in this case is custom CNN model. When the framework is applied on the Vision System, non-robust outputs corresponding to each models are recorded. Table~\ref{table:vis-sys} shows the erroneous predictions made by each model on the augmented test dataset corresponding to their causative corruption type. 
\begin{table*}[h]
\centering
\begin{tabular}{l|cccccccccc}
\hline
Models $f_i$ & \rotatebox{90}{shot\_noise} & \rotatebox{90}{impulse\_noise} & \rotatebox{90}{defocus\_blur} & \rotatebox{90}{glass\_blur} & \rotatebox{90}{motion\_blur} & \rotatebox{90}{snow} & \rotatebox{90}{frost} & \rotatebox{90}{fog} & \rotatebox{90}{brightness} & \rotatebox{90}{contrast} \\
\hline
yolov4-tiny ($f_0$) & 0.13 & 0.07 & 0.39 & 0.36 & 0.41 & 0.22 & 0.34 & 0.45 & 0.50 & 0.31 \\
speed\_sign ($f_3$) & 0.07 & 0.02 & 0.23 & 0.29 & 0.26  & 0.13 & 0.25 & 0.30 & 0.35 & 0.25 \\
main\_signal ($f_1$) & 0.05 & 0.013 & 0.22 & 0.21 & 0.17 & 0.03 & 0.07 & 0.10 & 0.20 & 0.06 \\
\hline
\end{tabular}
\caption{$\alpha_i$ attribution scores for Vision System generated by SETA}\label{table:vis-sys}
\end{table*}

\subsection{Ensemble Model}
\label{sec:ensemble}
As part of the experimental evaluation, an \textit{ensemble} of heterogeneous convolutional neural network (CNN) classifiers~\cite{resnet,vgg16} has been employed as a sanity check for the framework.
Ensemble models aggregate predictions from multiple independent learners, thereby improving average accuracy through diverse feature representations without introducing inter-dependencies between constituent models.
\begin{table}[h]
\centering
\begin{tabular}{c c c}
\toprule
 & Predicted Accuracy & Actual Accuracy \\
\midrule
ResNet18 &  84.94 & 83.09 \\
VGG16 & 81.74 & 79.84 \\
CustomCNN & 81.33  &  80.00 \\
\midrule
\multicolumn{2}{r}{\textbf{Ensemble}} & 86.63 \\
\bottomrule
\end{tabular}
\caption{Approximate accuracy values for each classifier model $f_i$ of an ensemble system as predicted by SETA.}\label{table:ensemble}
\end{table}
\noindent Each model in the ensemble was retrained on the CIFAR-10~\cite{cifar10} dataset, with the final fully connected layers modified to suit the classification task.
The custom CNN model consists of three convolutional blocks (each with convolution, ReLU activation, and max-pooling), followed by two fully connected layers, totaling over 1.1 million learnable parameters, and trained using Dropout~\cite{dropout} and the Adam optimizer~\cite{adamOptim}.
During evaluation, the framework tracks the number of times each model is invoked, as well as the occurrence of non-robust outputs under controlled perturbations.
This enables the calculation of model-specific and ensemble-level robustness statistics, effectively validating the accuracy of the fault attribution mechanism.
The empirical results, including individual model accuracies and overall ensemble performance, are summarized in Table~\ref{table:ensemble}.
By verifying consistency between conventional accuracy metrics and the framework’s recorded outcomes, we demonstrate the correctness and reliability of SETA in a multi-component vision-based system.

\subsection{OCR Model}
\label{sec:ocr}
To illustrate the generality of SETA beyond purely vision-based pipelines, the framework was also applied to an optical character recognition (OCR) system comprised of \textit{EasyOCR} and similar models.
The system was evaluated on multiple natural scene OCR datasets~\cite{ShopSign,UberText,IIIT5K}, with perturbations sampled from \textit{img\_aug}~\cite{imgaug}, \textit{albumentations}~\cite{albumentations}, and \textit{imagecorruptions}~\cite{michaelis_imagecorruptions_2020} libraries to simulate real-world image distortions, meteorological phenomena, sensor noise, etc.

In this setup, an OCR module $f_\textit{ocr}(x)$ produces a predicted text sequence $y$ for any given input image $x$.
The input is then transformed under some perturbation $g(x)$ sampled from one of the libraries mentioned above and returns a new perturbed input-output pair $\tilde{x}$, $\tilde{y}$, such that, $f_\textit{ocr}(\tilde{x}) = \tilde{y}$.
Output consistency is then quantified by comparing the predictions on the original and perturbed images using \textit{Levenshtein distance}.
A module is considered consistent if the Levenshtein distance between its original and perturbed outputs is below a threshold $\tau$, i.e., $d(y, \tilde{y}) \leq \tau$.
Therefore, the metamorphic relation assigned to this particular task is $M_\textit{ocr}(y, \tilde{y}; f_\textit{ocr}) \coloneqq H(\theta, \tau)$, where $\theta = - d(y, \tilde{y})$ is the Levenshtein distance between the predicted sequence for original and perturbed examples and the threshold is of the form $\tau \in \mathbb{Z}_{\leq 0}$.
This approach effectively adapts the SETA workflow for robustness benchmarking in text-based systems, recording module activations, deviations, and computing failure contribution scores in the same manner as for vision pipelines demonstrated above.

\begin{table}[htbp]
\centering
\begin{tabular}{c c c}
\toprule
Input ID & Robustness & Avg. Levenshtein Distance \\
\midrule
0 & 0.000 & 74.81 \\
1 & 1.000 & 0.00 \\
2 & 0.062 & 9.44 \\
3 & 1.000 & 0.63 \\
4 & 1.000 & 0.12 \\
5 & 0.375 & 2.63 \\
6 & 0.812 & 1.69 \\
\midrule
\multicolumn{2}{r}{\textbf{Average robustness}} & 0.607 \\
\bottomrule
\end{tabular}
\caption{OCR robustness evaluation under perturbations ($-\tau = 2$).}\label{table:ocr-robustness}
\end{table}
\noindent SETA can systematically track component-level anomalies and attribute failures in OCR systems, highlighting the framework’s extensibility to diverse, multi-component AI applications where output consistency under perturbation is a critical property.
Robustness in this setup indicates the fraction of perturbations for which the predicted text is within Levenshtein distance threshold

In the context of the SETA framework, the robustness values in Table~\ref{table:ocr-robustness} serves as module-level composite metamorphic relation scores $S_i(y, \tilde{y})$ for each OCR component.
Inputs with robustness below $1.0$ indicate deviations from expected behaviour under perturbations and contribute to the module’s failure count in the computation of $\alpha_i$.
By aggregating these deviations across the dataset, SETA quantifies the relative influence of each OCR module on system-level inconsistencies, enabling a data-driven prioritization of components for inspection or retraining.
Thus, even in a text-based AI system, the combination of Levenshtein distance thresholds and execution trace analysis allows systematic identification of the modules most responsible for non-robust outputs, demonstrating the extensibility of SETA beyond purely vision-based pipelines.

\section{Experimental Results}
\label{sec:results}

Figure~\ref{fig:tree} illustrates how different modules in the system exhibit varying sensitivities to diverse image perturbations.
For instance, the object detection module $f_0$ is particularly vulnerable to \texttt{fog}, whereas the image classification models $f_1 \dots f_5$ demonstrate greater robustness in their respective tasks.

During the experiments, only modules $f_1$ and $f_3$ were actively invoked to generate predictions, and accordingly, fault attribution scores ($\alpha_i$) were computed for only these modules.
Modules that were not triggered by any reference or perturbed inputs remain at their default $\alpha_i = 0$, reflecting the framework’s conservative design in avoiding spurious attributions.
The confidence in the assigned $\alpha_i$ values increases proportionally with the frequency of a module’s activation, as more observations strengthen the statistical association between deviations in module outputs and system-level failures.

These attribution scores quantify correlations between module-level inconsistencies and end-to-end system failures.
When strong and sufficiently comprehensive metamorphic relations are defined for each module, these correlations can approximate causal influences, providing principled insights into fault propagation.
Achieving this, however, requires carefully designing metamorphic relations that capture the specific anomalous behaviours of interest.
With appropriate access to module internals, SETA enables the formulation of high-fidelity metamorphic relations, allowing analysts to reason about how errors propagate through the system, influence downstream computations, and affect both predictions and routing mechanisms.

Importantly, the applicability of SETA extends beyond classical vision pipelines.
In the ensemble model experiments, module-level attribution allowed us to identify which classifiers in the ensemble contributed most to output inconsistencies, providing a sanity check on the framework’s statistical reasoning.
Similarly, in the OCR experiments using the \textit{EasyOCR} model, SETA quantified robustness under perturbations on the multiple natural scene OCR datasets by comparing Levenshtein distances of predicted text sequences.
Here, low robustness scores flagged modules whose predictions deviated beyond a threshold ($-\tau = 2$) under input transformations, demonstrating that the same principles of execution trace analysis and metamorphic relation-based scoring apply seamlessly to text-output systems.
Together, these evaluations illustrate the generality of SETA across heterogeneous AI pipelines, highlighting its ability to diagnose and attribute faults in both vision- and text-based multi-component systems.

\section{Conclusion}
\label{sec:conclusions}

This paper presents \textit{State-based Execution Trace Analysis (SETA)}, a proof-of-concept framework for diagnosing emergent failures in compound AI systems composed of interacting neural modules.
SETA addresses a critical limitation in the testing of such systems, namely, the inability of conventional end-to-end evaluation to isolate the source of failures in multi-component pipelines.
By combining principles from \textit{distributed tracing} and \textit{Metamorphic Testing (MT)}, SETA reconstructs per-input execution trees, monitors deviations in component-level behavioural relations, and statistically attributes system-level anomalies to the modules most likely responsible.
Unlike traditional testing strategies that rely on explicit ground-truth oracles, SETA leverages metamorphic relations as self-consistency specifications, enabling automated, pseudo-oracle validation at scale.
The proposed framework’s key contribution lies in its statistical fault attribution mechanism, which aggregates observed deviations across perturbed executions to compute a normalized \textit{Failure Contribution Score} for each module.
This score quantifies a component’s empirical influence on observed system failures, providing a principled diagnostic signal that guides engineers toward vulnerable or unreliable submodules.
Preliminary validation on a safety-critical computer vision pipeline demonstrates the feasibility of the approach and its potential to uncover subtle module-level weaknesses that propagate to system-level faults.

Nevertheless, SETA remains an early-stage effort with several open challenges.
The current formulation provides correlational, not causal, fault attribution; high module scores indicate association with failure rather than definitive responsibility.
The diagnostic precision of the method also depends on the coverage and expressiveness of the metamorphic relations used, which are presently hand-crafted and may be incomplete.
Future research should explore integrating causal inference or interventional analysis to move from correlation toward causal fault localization, and leveraging program synthesis or learning-based methods to automatically infer metamorphic relations from data and execution logs.
Further, extending SETA beyond vision pipelines to multimodal or reinforcement-learning systems could generalize its applicability to broader classes of AI-driven software.
Ultimately, this work establishes a foundation for compositional testing and explainable fault localization in complex AI systems, positioning SETA as a stepping stone toward systematic, interpretable, and causally grounded approaches for ensuring AI system reliability.

\begin{acks}
To Dr. Girish Maskeri Rama and Dr. Snigdha Athaiya for their gracious support and feedback throughout the process.
\end{acks}

\bibliographystyle{ACM-Reference-Format}
\bibliography{bibliography}



\end{document}